\pdfoutput=1

\documentclass[11pt]{article}

\usepackage[]{acl}

\usepackage{times}
\usepackage{latexsym}

\usepackage[T1]{fontenc}

\usepackage[utf8]{inputenc}

\usepackage{microtype}

\usepackage{pgfplots}
\pgfplotsset{
    width=7cm,
    /tikz/mark size=1.5pt,
    /tikz/mark=*,
    /tikz/mark options={solid},
    scaled x ticks=false,
    compat=1.9
}

\usepackage{enumitem}

\usepackage{stfloats}

\usepackage{array}

\usepackage{xcolor}

\usepackage{soul}

\usepackage{CJK}
\newcommand{\cjk}[1]{\begin{CJK}{UTF8}{gbsn}#1\end{CJK}}

\usepackage{subfigure}

\usepackage{multirow}

\title{PETCI: A Parallel English Translation Dataset of Chinese Idioms}

\author{Kenan Tang \\
  The University of Chicago \\
  \texttt{kenantang@uchicago.edu} \\}

\begin{document}

\maketitle

\thispagestyle{plain}
\pagestyle{plain}

\begin{abstract}
Idioms are an important language phenomenon in Chinese, but idiom translation is notoriously hard. Current machine translation models perform poorly on idiom translation, while idioms are sparse in many translation datasets. We present \textbf{PETCI}, a parallel English translation dataset of Chinese idioms, aiming to improve idiom translation by both human and machine. The dataset is built by leveraging human and machine effort. Baseline generation models show unsatisfactory abilities to improve translation, but structure-aware classification models show good performance on distinguishing good translations. Furthermore, the size of PETCI can be easily increased without expertise. Overall, PETCI can be helpful to language learners and machine translation systems.
\end{abstract}

\section{Introduction}
Idioms are an important yet complex language phenomenon. In certain corpora, 3 out of 10 sentences are estimated to contain idioms \citep{moon1998fixed}. The pervasively usage of idioms entails a wide range of linguistic studies \citep{cacciari2014idioms}. In NLP, idioms are involved in various tasks, including idiom recognition \citep{peng-feldman-2016-experiments, liu-hwa-2018-heuristically}, embedding learning \citep{tan-jiang-2021-learning}, and idiom comprehension \citep{zheng-etal-2019-chid}.

However, these NLP systems assume a monolingual setting, and the models perform poorly due to the non-compositional nature of idioms. Adopting a multilingual setting may be beneficial, but existing multilingual idiom datasets are rather small \citep{moussallem-etal-2018-lidioms}. Machine translation systems also cannot help, because idiom translation is challenging \citep{salton-etal-2014-evaluation, cap-etal-2015-account, fadaee-etal-2018-examining}. Idiom translation is even difficult for human, as it requires knowing both the source and target culture and mastering diverse translation strategies \citep{wang2013study}. 

Despite these difficulties, there are not enough datasets for model improvement. Translating idioms from scratch is very laborious. Nevertheless, existing idiom dictionaries have always been readily available. Moreover, they usually provide more than one translation to each idiom, which naturally results in a paraphrasal dataset. The paraphrase strategies used for idiom translation are usually more aggressive than in other paraphrasal datasets such as PPDB \citep{ganitkevitch-etal-2013-ppdb}. Therefore, neural network models based on these dictionaries may benefit idiom translation more than naive retrieval.

To this end, we present \textbf{PETCI}, a \textbf{P}arallel \textbf{E}nglish \textbf{T}ranslation dataset of \textbf{C}hinese \textbf{I}dioms. PETCI is collected from an idiom dictionary, together with machine translation results from Google and DeepL. Instead of directly using models to translate idioms, we design tasks that require the model to distinguish gold translations from unsatisfactory ones and to rewrite translations to improve their quality. We show that models perform better with the increase of the dataset size, while such increase requires no linguistic expertise. PETCI is publicly available\footnote{\url{https://github.com/kt2k01/petci}}.

The main contribution of this paper is two-fold. First, we collect a dataset that aims improve Chinese idiom translation by both machine translation systems and language learners. Secondly, we test several baseline models on the task we defined.

In the following sections, we begin by a detailed description of our data collection process (Section \ref{section:data-collection}), followed by case studies and statistics (Section \ref{section:statistics}). Based on the dataset, we define tasks and report the performance of baseline models we choose (Section \ref{section:model}, \ref{section:experiment}, \ref{section:results}). Then, we discuss our observations, provide relevant work and future directions (Section \ref{section:related-work}), and conclude (Section \ref{section:conclusion}). 

\begin{table*}[t]
\centering
\begin{tabular}{c|p{4cm}|p{4cm}|c}
\hline
\textbf{Chinese} & \multicolumn{1}{c|}{\textbf{Dictionary Translation}} & \multicolumn{1}{c|}{\textbf{Machine Translation}} & \textbf{Issues Identified}\\
\hline
\cjk{为虎傅翼}
&
\begin{tabular}{@{\textbullet~}p{3.6cm}@{}}
    assist an evil-doer \\
    give wings to a tiger \\
    increase the power of a tyrant to do evil \\
    \textcolor{red}{\ul{lend support to an evil-doer like adding wings to a tiger}}
\end{tabular}
&
\begin{tabular}{@{\textbullet~}p{3.6cm}@{}}
    \textcolor{blue}{\ul{Fu Yi}} for the tiger \\
    for the tiger's \textcolor{magenta}{\ul{wings}} \\
    for the tiger \\
    for the tiger's \textcolor{magenta}{\ul{wing}} \\
    for the tiger's \textcolor{magenta}{\ul{sake}}
\end{tabular}
&
\begin{tabular}{c}
    \textcolor{blue}{Name hallucination} \\
    \textcolor{magenta}{Single-word edit} \\
    \textcolor{red}{Very long alternative}
\end{tabular}
\\ \hline
\cjk{煮豆燃萁}
&
\begin{tabular}{@{\textbullet~}p{3.6cm}@{}}
    fratricidal strife \\
    burn beanstalks to cook beans\textcolor{magenta}{\ul{ - }}one member of a family injuring another \\
    boil beans with bean-stalks\textcolor{magenta}{\ul{ - }}reference to a fight among brothers \\
\end{tabular}
&
\begin{tabular}{@{\textbullet~}p{3.6cm}@{}}
    \textcolor{blue}{\ul{boiled beans}} \\
    burning beanstalks cook the beans (idiom)\textcolor{red}{\ul{; }}to cause internecine strife \\
\end{tabular}
&
\begin{tabular}{c}
    \textcolor{blue}{Partial translation}\\
    \textcolor{magenta}{Hyphenation}\\
    \textcolor{red}{Semicolon}
\end{tabular}
\\
\hline
\cjk{风流云散}
&
\begin{tabular}{@{\textbullet~}p{3.6cm}@{}}
    \textcolor{blue}{\ul{(of old companions)}} separated \textcolor{red}{\ul{and}} scattered\\
    blown apart by the wind \textcolor{red}{\ul{and}} scattered like clouds\\
    \textcolor{blue}{\ul{(of relatives or friends, separated from each other)}} as the wind blowing \textcolor{red}{\ul{and}} the clouds scattering\\
\end{tabular}
&
\begin{tabular}{@{\textbullet~}p{3.6cm}@{}}
    \textcolor{magenta}{\ul{wind and clouds}}\\
    \textcolor{magenta}{\ul{Wind flow and clouds scatter}}\\
    \textcolor{magenta}{\ul{Wind flow and clouds scattered}}\\
\end{tabular}
&
\begin{tabular}{c}
    \textcolor{blue}{Parenthesis} \\
    \textcolor{magenta}{Literal translation} \\
    \textcolor{red}{Explicit parallelism}
\end{tabular}
\\ \hline
\cjk{萧规曹随}
&
\begin{tabular}{@{\textbullet~}p{3.6cm}@{}}
    follow established rules\\
    \textcolor{blue}{\ul{Tsao}} (a Han Dynasty prime minister) followed the rules set by \textcolor{blue}{\ul{Hsiao}} (his predecessor)\\
\end{tabular}
&
\begin{tabular}{@{\textbullet~}p{3.6cm}@{}}
    \textcolor{blue}{\ul{Xiao}} Gui \textcolor{blue}{\ul{Cao}} Sui\\
    \textcolor{magenta}{\ul{lit. Xiao Gui Cao}} follows the rules \textcolor{magenta}{\ul{(idiom); fig.}} follow the rules of a certain place\\
\end{tabular}
&
\begin{tabular}{c}
    \textcolor{blue}{Different romanization} \\
    \textcolor{magenta}{Unstable reference}
\end{tabular}
\\ \hline
\end{tabular}
\caption{Examples of idioms and translations in PETCI. Parts of the translations are colored and underlined, if they relate to the identified issues (details in Section \ref{section:case-study}). The first Machine translation is always from Google.}
\label{tab:case-study}
\end{table*}

\section{Data Collection}
\label{section:data-collection}
\subsection{Human Translation}
We collect human translations from a English translation dictionary of Chinese idioms \citep{liu_1993}. This dictionary is preferred because it provides multiple translations to a single idiom, with one of them labelled as being most frequently used. Though the dictionary is published 3 decades ago, the time is short compared to the observed long time for the meaning of idioms to change \citep{nunberg1994idioms}. 

We use the OCR tool \texttt{tesseract}\footnote{\url{https://tesseract-ocr.github.io/}} to extract the English translations from a scanned version of the book. The Chinese idioms are then manually added. We performed both automatic and manual check to ensure the quality of final results (details in Appendix \ref{section:cleaning}).

\subsection{Machine Translation}
We collect machine translations from the publicly available models of Google and DeepL\footnote{We do not use the Baidu translation that has better performance on Chinese, because it directly returns a dictionary translation for most of the time.}. Though they both provide free APIs for developers, richer information is returned from direct queries on the website interface (details in Appendix \ref{section:interface}). For example, while Google always returns a single translation, DeepL sometimes returns alternative translations.

We decide to use the alternative translations. Therefore, instead of using developer APIs, we have to scrape translations from the web interface. Because the shadow DOM prevents the content from being scraped by \texttt{selenium}, we use a script that automatically paste our source idiom into the source box, and take a screenshot of the result box. The \texttt{tesseract} OCR tool is then used on the screenshots to retrieve the text. Though OCR performs much better on screenshots than on scans as expected, we manually check the results to ensure quality.

\subsection{Case Study}
\label{section:case-study}

Here, we summarize our observations during data collection. Examples are listed in Table \ref{tab:case-study}. In order to avoid confusion, for the rest of this paper, we use \textbf{Gold} translations to refer to the set of the first translation of all idioms in the \textbf{Dictionary} translations. The rest of the Dictionary translation is denoted as \textbf{Human} translations. We use \textbf{Machine} translations to refer the combination of \textbf{Google} and \textbf{DeepL} translations. In other words, we have:
\begin{itemize}
    \item $\textrm{Dictionary} = \textrm{Gold} \cup \textrm{Human}$
    \item $\textrm{Machine} = \textrm{Google} \cup \textrm{DeepL}$
\end{itemize}

\begin{description}[style=unboxed,leftmargin=0cm, labelsep=1em, itemsep=0em]
    \item[Non-uniform Numbers of Alternatives] Alternative translations are provided by both the Dictionary and DeepL. The number of alternatives from the Dictionary varies from 0 to 13. DeepL provides only up to 3 alternatives, but the first translation sometimes contains two translations separated by a semicolon.
    \item[Low Edit Distance] The multiple alternatives from DeepL usually result from single-word edits, when the translation itself is much longer than one word. This almost never happens in the Dictionary. 
    \item[Dictionary Artifacts] The Dictionary translations aim at human readers, so part of the translation may only be explanatory. Indicators of explanation include parenthesis, dash, and abbreviations such as \emph{sb.} (somebody) and \emph{sth.} (something). We expand the abbreviations but keep the other explanations. Also, due to the time it was published, the Dictionary uses the Wade-Giles romanization system, which is different from the Hanyu Pinyin used by Machine. 
    \item[Machine Artifacts] Some Machine translations can be the same as Human or even Gold ones. However, mistakes do exist. We notice meaningless repetition and spurious uppercase letters. In some cases, the uppercase letters seems to result from the machine recognizing part of an idioms as names. In very rare cases, Machine translations contain untranslated Chinese characters. We substitute them by the Hanyu Pinyin. All these artifacts are unique to Machine translations.
    \item[Unstable References] DeepL sometimes refers to other dictionaries. However, DeepL may replace part of the correct dictionary translation by machine artifacts. Also, DeepL uses abbreviations inconsistently, indicating that it refers to multiple dictionaries. Despite the lack of citation information, we identify one of the dictionaries to be \emph{4327 Chinese idioms} \citep{akenos_2012}, which uses the two abbreviations \emph{fig.} and \emph{lit.}.
    \item[Literal Machine Translation] The literal Machine translation is even less satisfactory than the literal Human translation for two reasons. First, Machine sometimes returns a partial translation of only 2 out of the 4 Chinese characters. Secondly, Machine may translate the characters one by one, ignoring the non-compositionality of idioms.
    \item[Explicit Structure] Parallel structures are pervasive in Chinese idioms \citep{wu1995cultural} but usually not explicitly indicated by a conjunction character. Parallelism can be revealed in the English translation by conjunctions. However, parallel structures are sometimes considered redundant, and abridging is used to remove parallelism in translations of Chinese idioms without the loss in meaning \citep{wang2013study}. Structures other than parallelism are discussed in \citep{wang-yu-2010-construction}, but they cannot be naively detected from translations.
    \item[Optimality of Gold Translations] Many Human translations are more verbose than the Gold translation. This is consistent with the tendency of language learners to err on the verbose and literal side. However, many Human alternatives are also much more succinct than the Gold translation, which can be arguably better. Without further linguistic analysis, we accept the Gold translations as labelled by the dictionary.
\end{description}

\section{Statistics}
\label{section:statistics}

In this section, we compute statistics of PETCI, and compare with the existing idiom datasets. We then use our dataset to probe the appearance of idioms in other translation datasets. We also provide a quantitative analysis for some our observations in Section \ref{section:case-study}.
\subsection{Comparison with Existing Datasets}
We summarize our comparison of dataset sizes in Table \ref{tab:data-size}. Monolingual idiom dictionaries in both English and Chinese are usually much larger than bilingual idiom dictionaries. For comparison, we choose \emph{Oxford Dictionary of Idioms} \citep{ayto2020oxford}, \emph{Xinhua Idiom Dictionary} \citep{xinhua}, and \emph{4327 Chinese idioms} \citep{akenos_2012}. The precise number of translations in \emph{4327 Chinese idioms} is not available, but the dictionary usually provides both a literal and a figurative translation. Therefore, we estimate the number of translations to be twice the number of Chinese idioms.

There have also been attempts to build a dataset of English translated Chinese idioms. For example, the CIBB dataset contains the English translation of Chinese idiom, together with a blacklist of words that indicate unwanted literal translation \citep{shao-etal-2018-evaluating}. This dataset has been used to score translation models \citep{huang-etal-2021-comparison}, but it has an extremely small size of only 50 distinct Chinese idioms and a pair of translations for each idiom. Another dataset, CIKB, thoroughly contains 38,117 idioms with multiple properties \citep{wang-yu-2010-construction}. Among all properties, each idiom has at most 3 English translations, including the literal translation, free translation, and the English equivalent. However, of the 38,117 idioms, only 11,000 have complete properties (among which 3,000 are labelled as most commonly used). Though the CIKB dataset is not made public, we reasonably assume that the missing properties are mainly English translations, because the other properties are in Chinese and thus easier to obtain. We estimate the number of English translations in CIKB to be between 3 times the number of complete entries and of all entries.

\begin{table}
\centering
\begin{tabular}{ccc}
\hline
\textbf{Dataset} & \textbf{Chinese} & \textbf{English}\\
\hline
\emph{Oxford} & --- & 10,000 \\
\emph{Xinhua} & 10,481 & ---\\
\emph{4327} & 4,327 & 8,654 \\ 
CIBB & 50 & 100 \\ 
CIKB & 38,117 & 33K - 114K \\
PETCI & 4,310 & 29,936 \\
\hline
\end{tabular}
\caption{The comparison between the sizes of dictionaries and datasets. Some sizes are estimated.}
\label{tab:data-size}
\end{table}

From our comparison, we see that PETCI covers a large percentage of frequently used Chinese idioms, while uniquely providing much more parallel translations. To further grow the size of PETCI, we propose 3 orthogonal directions: (1) increase the number of idioms, (2) add more translations to each idiom, and (3) add multiple gold translations to each idiom. The first two directions can be accomplished by language learners without linguistic expertise. The third direction, though requiring expertise, is revealed to be useful by the model performance.

\subsection{Percentage of Chinese Idioms}
We would like to know how many sentences in widely used Zh-En datasets contain at least one idiom from our dataset. We consider the datasets from the news translation task from WMT21, with intuitively the highest percentage of idiom usage among all provided domains. The result in Table \ref{tab:proportion} shows that all the datasets have a low Percentage of sentences using Chinese Idioms (PoI), with the PoI of the two datasets related to Wikipedia lower than 1\%.

\begin{table}
\centering
\begin{tabular}{lcc}
\hline
\textbf{Dataset} & \textbf{Size (M)} & \textbf{PoI (\%)}\\
\hline
ParaCrawl v7.1 & 14.17 & 1.56 \\
News Commentary v16 & 0.32 & 7.70\\
Wiki Titles v3 & 0.92 & 0.02 \\ 
WikiMatrix & 2.60 & 0.90\\ 
Back-translated news & 19.76 & 1.75\\\hline
\end{tabular}
\caption{The Percentage of sentences using Chinese Idioms (PoI) of different datasets from the WMT21 news translation task.}
\label{tab:proportion}
\end{table}

The low PoI is not only observed in the Zh-En pair \citep{fadaee-etal-2018-examining}. Also, the PoI might differ in the En-Zh and Zh-En directions, though the available Zh-En datasets from WMT are all undirectional. 

\subsection{Quantitative Case Study}
\label{section:quant-case-study}

In this subsection, we qualitatively analyze our observations in the case study (Section \ref{section:case-study}). The results are summarized in Table \ref{tab:qual}. 

\begin{table*}[t]
\centering
\begin{tabular}{l|cccccc}
\hline
\textbf{Statistics} & \textbf{Dictionary} & \textbf{Gold} & \textbf{Human} & \textbf{Machine} & \textbf{Google} & \textbf{DeepL}\\
\hline
Total number & 14997 & 4310 & 10687 & 14939 & 4310 & 10629\\
Avg. length (token) & 5.03 & 4.97 & 5.06 & 3.72 & 2.81 & 4.09\\
Avg. length (char) & 27.14 & 27.15 & 27.14 & 19.99 & 15.72 & 21.72\\
\hline
\% longer (token) & --- & --- & 47.33 & 25.44 & 11.39 & 34.14\\
\% shorter (token) & --- & --- & 33.16 & 54.75 & 71.07 & 48.13\\
\% longer (char) & --- & --- & 56.04 & 30.05 & 14.80 & 36.23\\
\% shorter (char) & --- & --- & 40.57 & 63.82 & 78.52 & 57.86\\
\hline
\% with NNP & 3.67 & 0.77 & 2.99 & 4.43 & 2.99 & 2.09\\
\% with CC & 39.72 & 18.26 & 31.00 & 32.44 & 9.84 & 30.32\\
\hline
Avg. edit distance & 5.24 & --- & --- & --- & --- & 2.45 \\
\% of single edit & 0.83 & --- & --- & --- & --- & 27.20 \\
\hline
\end{tabular}
\caption{The statistics of different subsets in PETCI. \emph{Longer} and \emph{shorter} are based on comparison with the Gold translations, therefore not applicable to Dictionary translations. \emph{Edit distances} are only available for subsets that contain multiple translations. Human is a subset of Dictionary, so its average edit distance is not calculated.}
\label{tab:qual}
\end{table*}

\begin{description}[style=unboxed,leftmargin=0cm, labelsep=1em, itemsep=0em]
    \item[Length] The average length of Chinese idioms in PETCI is 4.30, with 91.42\% of all idioms having 4 characters. We tokenize translations using the \texttt{stanza} package \citep{qi-etal-2020-stanza}, and report the average length of translations in tokens and characters. Surprisingly, the length of Human translations and Gold translations are extremely close. By calculating the percentage of different translations that are longer or shorter than the gold translation, we see that long sub-optimal translations are still unique to Human translations. The Machine translations tend to be shorter due to partial translation.
    \item[Name Hallucination] We would like to know how much hallucinated names has been introduced in the Machine translations. We used \texttt{stanza} to perform part-of-speech (POS) tagging on the translations, and calculate the percentage of idioms that has at least one NNP token in its translation. The Machine translations have a NNP percentage much higher than the Gold translation, but close to the Human translations. This may result from that Human translations include explanations that involve names in a story from which a idiom originates. 
    \item[Parallel Structure] The popular Chinese word segmentation package \texttt{jieba} considers 93.64\% of the Chinese idioms in PETCI as an atomic unit. However, the idioms obviously have internal structures. For a coarse estimation, we consider a translation containing the CC token to have parallel structure. We calculate the percentage of idioms that has at least one CC token in its translations. We see that parallel structure is pervasive in PETCI, but less favored by the Gold translation. Also, the DeepL translations can catch parallel structures as well as the Human translations. Google performs less well.
    \item[Edit Distance] To quantitatively analyze the simple-word edit strategy used by DeepL, we calculate the average token-level Levenshtein distance of all pairs of translation for the same idiom. We then calculate the percentage of single-word edit pairs, with the length of translation being at least 2 tokens. The same calculation is also applied to Dictionary translations. We notice a lower edit distance and higher percentage of single-word edits from DeepL, which is consistent with our qualitative observation. This suggests the possibility of further augmenting the data by single-word edit methods, such as replacing synonyms based on WordNet \citep{miller-1992-wordnet}.
    \item[Split and Filter] We observe that DeepL may join multiple translations into one, and that all translations may sometimes include dictionary artifacts. Therefore, all statistics is performed after the removal of 2,134 parenthesized content and 341 abbreviations. To further prepare the dataset for training, we split 250 translations that include a semicolon. After the split, we remove translations from the Machine set that also appeared in the Dictionary. The processed Human/Machine translation set has a size of 10,690/13,539. In the following sections, we have:
    \begin{itemize}
        \item $\textrm{Machine} \leftarrow Split(\textrm{Machine}) \setminus \textrm{Dictionary}$
    \end{itemize}
    
\end{description}

Explicitly incorporating the above translation features into models may improve their performance, but this strategy is outside the scope of this paper. For all models, we only use the text as the input.

\section{Models}
\label{section:model}
A human expert is able to tell whether a translation is satisfactory, and then rewrite the translation if it is not. We expect the models to behave similarly. Therefore, we design a binary classification task and a rewriting task. The binary classifier should determine whether a given translation is Gold or not. The rewriting model should rewrite other translations in the standard of Gold ones. The two tasks may be combined in a pipeline, but we leave that for future work, and use the following models to perform either of the two tasks. No major changes are made on the model architectures, so we omit detailed descriptions and refer readers to the original papers.

\subsection{LSTM Classification}

The recurrent neural network (RNN) with LSTM cells \citep{hochreiter1997long} is able to encode a sentence of variable length into a hidden state vector. Then, the vector can be fed into a linear layer for classification.

Some variants of LSTM, such as Bidirectional LSTM and Multilayer LSTM \citep{6707742}, can perform better on longer sequences, but most translations in PETCI are short. Moreover, we would like to introduce fewer variables when comparing models, so we only use the one-directional single-layer LSTM.

\subsection{Tree-LSTM Classfication}
We have shown that the translation of Chinese idioms exhibits special structures that may be used to improve model performance. A variant of LSTM that explicitly accounts for tree structures is the Tree-LSTM \citep{tai-etal-2015-improved}.

The attention mechanism can be further incorporated into the Tree-LSTM \citep{8665673}, but we do not use it for the sake of comparison. 

\subsection{BERT Classification}
The BERT model has been reported to have superior performance on sentence classification tasks \citep{devlin-etal-2019-bert}. It can also capture the structural information of the input text \citep{jawahar-etal-2019-bert}.

We do not use better variants of BERT for the following considerations. First, some variants are pre-trained on different datasets other than the BookCorpus \citep{bookcorpus} and English Wikipedia used by BERT. RoBERTa, for example, has part of its pre-training data from Reddit and CommonCrawl \citep{liu2019roberta}. Intuitively, idioms are used more in the colloquial language of Reddit. Though we are not able to estimate the English PoI in these datasets, we do see that WikiMatrix has a much lower PoI than ParaCrawl. Similarly, we expect the Wikipedia dataset to have a lower PoI than CommonCrawl. The choice of pre-training datasets by RoBerta may benefit our downstream task, but not ideal for fair comparison.

Secondly, other variants of BERT may improve the sematic representation of phrases and sentences, such as PhraseBERT \citep{wang-etal-2021-phrase} and SentenceBERT \citep{reimers-gurevych-2019-sentence}. However, they are also not ideal, because idiom translation covers a large range of text length. A user of the model may prefer to give either over-simplified (like Machine) or over-extended (like Human) translations, and the model is expected to perform well in both cases. Therefore, we take the vanilla BERT as a middle ground.

\subsection{OpenNMT Rewriting}

Rewriting is a sequence-to-sequence task. We use the OpenNMT framework which provides an encoder-decoder model based on LSTM with the attention mechanism \citep{klein-etal-2017-opennmt}. A part of the rewriting task can be viewed as simplification, on which the OpenNMT framework is reported to perform well \citep{nisioi-etal-2017-exploring}.

We do not consider the SOTA sequence-to-sequence models, such as BART \citep{lewis-etal-2020-bart} or T5 \citep{raffel2019exploring}, because they are much larger and forbiddingly difficult for language learners to use. 

\section{Experiments}
\label{section:experiment}
\subsection{Classification}
For the binary classification, we use train/dev/test splits of 3448/431/431 idioms. To study the effect of blending Human and Machine translation, we construct 3 types of training sets that contain (1) both Gold and Human (H), (2) both Gold and Machine (M), and (3) all Gold, Human, and Machine translations (HM). The development sets are similarly constructed and used when training on the corresponding set. The test set is the combination of Gold, Human, and Machine translations, meaning that models trained on the HM set will be evaluated on Machine translations for zero-shot accuracy. To balance the training data, whenever we add a Human or Machine translation into the training set, we always add its corresponding Gold translation once. The development and test sets are not balanced.

We also want to examine whether increasing the size of the training set may improve model performance. We construct partial training sets from the full set. In the case of training set H, the smallest partial set is constructed by taking exactly one Human translation from each idiom in the training set, together with Gold translations. The size of 3 other partial sets are then evenly spaced between the sizes of the smallest and the full training set, resulting in 5 H sets in total. For the 3 partial sets with middle sizes, we randomly sample from the remaining Human translations to reach the target size. The sizes of 5 H training sets are 3,050/4,441/5,833/7,224/8,616, expressed in the number of non-Gold translations (the total size is strictly twice as large). We similarly construct 5 M sets (sizes 3,419/5,265/7,112/8,958/10,805), and merge the H and M sets to create 5 HM sets. Here, distributional shift is inevitably introduced, but this process mimics the real-world scenario where it is impossible for a model user to add extra Gold translations, but can only add their own non-Gold translations of an unknown distribution compared to the existing H and M sets.

For both LSTM and Tree-LSTM, we use the cased 300-dimensional GloVe vectors pre-trained on CommonCrawl \citep{pennington-etal-2014-glove} to initialize word representations. We allow the word representations to be updated during training to improve performance \citep{tai-etal-2015-improved}.

We use the binary constituency Tree-LSTM because it is reported to out-perform the dependency Tree-LSTM. We use CoreNLP to obtain the constituency parse trees for sentences \citep{manning-etal-2014-stanford}. The binary parse option (not available in \texttt{stanza}) is set to true to return the internal binary parse trees. Half-nodes with only 1 child are removed.

However, fine-grained annotation on each node is not available, so only full trees are used during training. All nodes other than the root have a dummy label 0. We make this modification on a PyTorch re-implementation of the Tree-LSTM \footnote{\url{https://github.com/dmlc/dgl/tree/master/examples/pytorch/tree_lstm}}.

We implement BERT classification using HuggingFace \citep{wolf2019huggingface}. The pre-trained model we use is the \texttt{bert-base-uncased}.
\subsection{Rewriting}
We use all Human and Machine translations for the idioms in the train split as the source training set. The corresponding Gold translations are the target. The development and test sets are similarly constructed from the dev/test splits. We do not change the training set size for rewriting, and this choice will be justified when we discuss the results.

Our model architecture is the same as the NTS model based on OpenNMT framework \citep{nisioi-etal-2017-exploring}. Though originally designed to perform text simplification, the NTS model has no component that explicitly performs simplification. We only change pre-trained embeddings from word2vec \citep{mikolov2013efficient} to GloVe.

\subsection{Hyperparameters and Training Details}

To ensure fair comparison, the hyperparameters for each model are fixed for different training sets. For LSTM and Tree-LSTM, we follow the choice of \citep{tai-etal-2015-improved}. For BERT, \citep{devlin-etal-2019-bert} suggest that a range of parameters work well for the task, so we choose a batch size of 16, a learning rate of 5e-5, and an epoch number of 3. For OpenNMT, we follow the configuration provided by \citep{nisioi-etal-2017-exploring}. More details can be found in Appendix \ref{section:training-details}.

\section{Results and Discussion}
\label{section:results}

\subsection{Classification}

The results are plotted in Figure \ref{fig:classify}, with numeric results in Appendix \ref{section:numeric}. The suffix of a model indicate the training set that it is trained on. 

\begin{description}[style=unboxed,leftmargin=0cm, labelsep=1em, itemsep=0em]

\item[Gold Sparsity] Due to the sparsity of gold translations, a model may blindly assign the gold label with low probability, and the accuracy is still high. Therefore, we compare all results to a random baseline that assigns the label based on the probability equal to the label frequency in PETCI. The accuracy of the random baseline is 75.35\% overall, 14.40\% on Gold, and 85.60\% on either Human or Machine. Most models underperform the random baseline. Also, the Gold accuracy of all models deteriorates with the increase in the training set size, but still above the random baseline. Our data balancing strategy does not seem to overcome the sparsity problem.

\item[Training Set Size] With the increase in training set size, the overall accuracy mostly increases. The increase in performance has not saturated, indicating the need of further increasing dataset size. The BERT-HM model trained on the most data performs best, closely followed by Tree-LSTM-HM. When trained on the most data, LSTM does not out-perform the random baseline. In few cases, the increase in dataset size hurts performance, such as for LSTM-M. This may be an artifact of high standard deviation.

\item[Human and Machine] The models trained either on H or M sets have a low zero-shot accuracy on the opposite side, while training on the merged HM sets benefits both. The LSTM-H model seems to a high accuracy comparable to the random baseline, even outperforming LSTM-M on Machine without ever seeing a Machine translation. However, this seems to be an artifact of over-assigning the non-Gold label. The good performance of -HM models indicates that it is reasonable to combine Machine and Human translation in the training set, even when the model is responsible for classifying translation that only comes from either MT systems or language learners. 

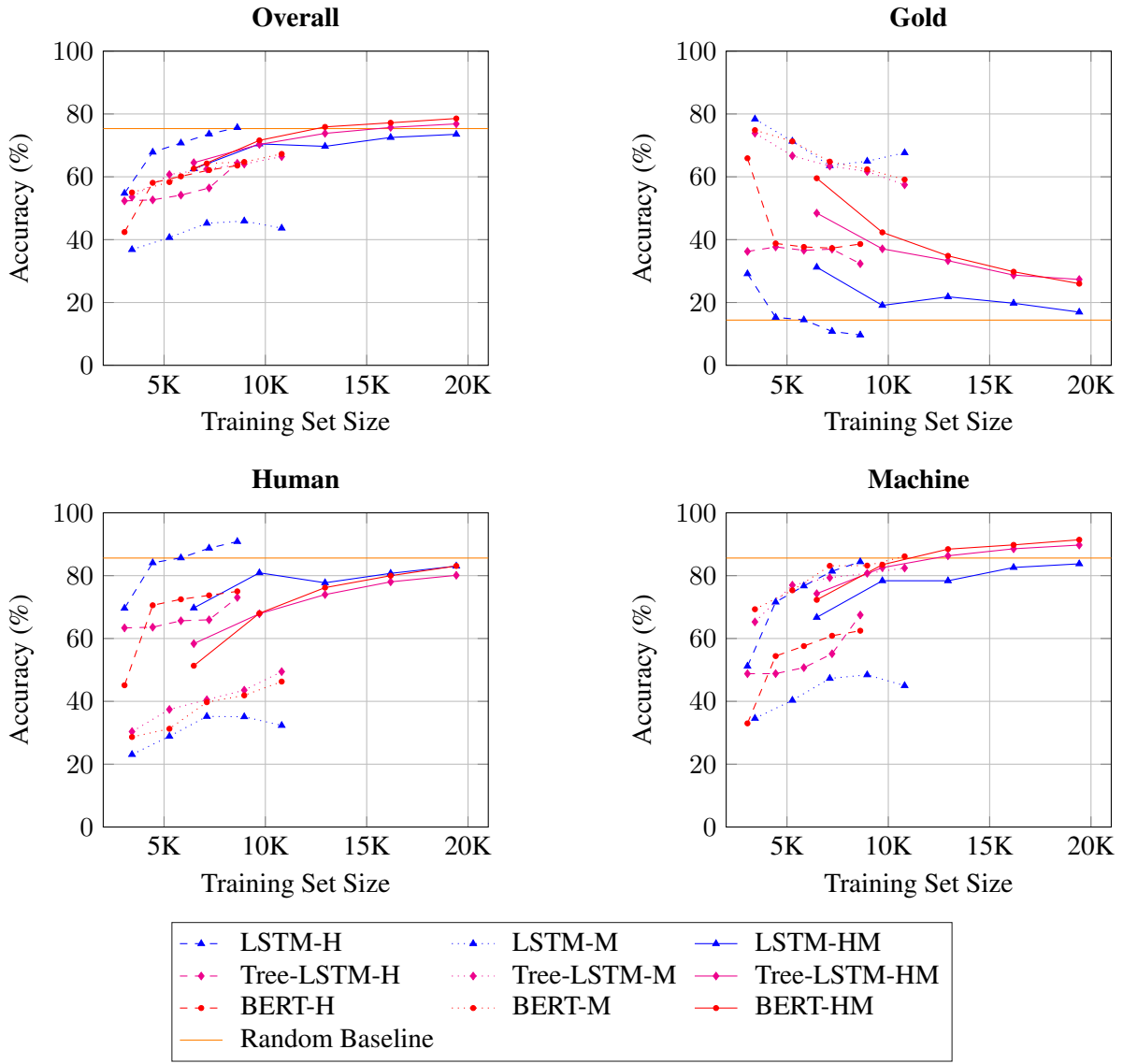
\begin{figure*}[t]
    \centering
    \begin{subfigure}{}
        \begin{tikzpicture}
            \begin{axis}[
                title={\textbf{Overall}},
                xlabel={Training Set Size},
                ylabel={Accuracy (\%)},
                xmin=2000, xmax=21000,
                ymin=0, ymax=100,
                xtick={5000, 10000, 15000, 20000},
                xticklabels={5K, 10K, 15K, 20K},
                legend pos=north west,
                ymajorgrids=true,
                xmajorgrids=true,
            ]
            \addplot [color=orange, no marks]
                table [x index=0, y index=1]
                {figs/accs/rnd-all.dat};
            
            \addplot [color=blue, mark=triangle*, dashed, mark options={solid}]
                table [x index=0, y index=1, y error index=2] {figs/accs/lstm-test-all-gh.dat};
            \addplot [color=blue, mark=triangle*, dotted, mark options={solid}]
                table [x index=0, y index=1, y error index=2] {figs/accs/lstm-test-all-gm.dat};
            \addplot [color=blue, mark=triangle*]
                table [x index=0, y index=1, y error index=2] {figs/accs/lstm-test-all-ghm.dat};
            \addplot [color=magenta, mark=diamond*, dashed, mark options={solid}]
                table [x index=0, y index=1, y error index=2] {figs/accs/tree_lstm-test-all-gh.dat};
            \addplot [color=magenta, mark=diamond*, dotted, mark options={solid}]
                table [x index=0, y index=1, y error index=2] {figs/accs/tree_lstm-test-all-gm.dat};
            \addplot [color=magenta, mark=diamond*]
                table [x index=0, y index=1, y error index=2] {figs/accs/tree_lstm-test-all-ghm.dat};
            \addplot [color=red, dashed, mark options={solid}, mark size=1pt]
                table [x index=0, y index=1, y error index=2] {figs/accs/bert-test-all-gh.dat};
            \addplot [color=red, dotted, mark options={solid}, mark size=1pt]
                table [x index=0, y index=1, y error index=2] {figs/accs/bert-test-all-gm.dat};
            \addplot [color=red, mark size=1pt]
                table [x index=0, y index=1, y error index=2] {figs/accs/bert-test-all-ghm.dat};
                
            \end{axis}
        \end{tikzpicture}
    \end{subfigure}
    \hfill
    \begin{subfigure}{}
        \begin{tikzpicture}
            \begin{axis}[
                title={\textbf{Gold}},
                xlabel={Training Set Size},
                ylabel={Accuracy (\%)},
                xmin=2000, xmax=21000,
                ymin=0, ymax=100,
                xtick={5000, 10000, 15000, 20000},
                xticklabels={5K, 10K, 15K, 20K},
                legend pos=north west,
                ymajorgrids=true,
                xmajorgrids=true,
            ]
            
            \addplot [color=orange, no marks]
                table [x index=0, y index=1]
                {figs/accs/rnd-g.dat};
            
            \addplot [color=blue, mark=triangle*, dashed, mark options={solid}]
                table [x index=0, y index=1, y error index=2] {figs/accs/lstm-test-g-gh.dat};
            \addplot [color=blue, mark=triangle*, dotted, mark options={solid}]
                table [x index=0, y index=1, y error index=2] {figs/accs/lstm-test-g-gm.dat};
            \addplot [color=blue, mark=triangle*]
                table [x index=0, y index=1, y error index=2] {figs/accs/lstm-test-g-ghm.dat};
            \addplot [color=magenta, mark=diamond*, dashed, mark options={solid}]
                table [x index=0, y index=1, y error index=2] {figs/accs/tree_lstm-test-g-gh.dat};
            \addplot [color=magenta, mark=diamond*, dotted, mark options={solid}]
                table [x index=0, y index=1, y error index=2] {figs/accs/tree_lstm-test-g-gm.dat};
            \addplot [color=magenta, mark=diamond*]
                table [x index=0, y index=1, y error index=2] {figs/accs/tree_lstm-test-g-ghm.dat};
            \addplot [color=red, dashed, mark options={solid}, mark size=1pt]
                table [x index=0, y index=1, y error index=2] {figs/accs/bert-test-g-gh.dat};
            \addplot [color=red, dotted, mark options={solid}, mark size=1pt]
                table [x index=0, y index=1, y error index=2] {figs/accs/bert-test-g-gm.dat};
            \addplot [color=red, mark size=1pt]
                table [x index=0, y index=1, y error index=2] {figs/accs/bert-test-g-ghm.dat};
                
            \end{axis}
        \end{tikzpicture}
    \end{subfigure}
    
    \begin{subfigure}{}
        \begin{tikzpicture}
            \begin{axis}[
                title={\textbf{Human}},
                xlabel={Training Set Size},
                ylabel={Accuracy (\%)},
                xmin=2000, xmax=21000,
                ymin=0, ymax=100,
                xtick={5000, 10000, 15000, 20000},
                xticklabels={5K, 10K, 15K, 20K},
                legend pos=north west,
                ymajorgrids=true,
                xmajorgrids=true,
            ]
            \addplot [color=orange, no marks]
                table [x index=0, y index=1]
                {figs/accs/rnd-hm.dat};
            
            \addplot [color=blue, mark=triangle*, dashed, mark options={solid}]
                table [x index=0, y index=1, y error index=2] {figs/accs/lstm-test-h-gh.dat};
            \addplot [color=blue, mark=triangle*, dotted, mark options={solid}]
                table [x index=0, y index=1, y error index=2] {figs/accs/lstm-test-h-gm.dat};
            \addplot [color=blue, mark=triangle*]
                table [x index=0, y index=1, y error index=2] {figs/accs/lstm-test-h-ghm.dat};
            \addplot [color=magenta, mark=diamond*, dashed, mark options={solid}]
                table [x index=0, y index=1, y error index=2] {figs/accs/tree_lstm-test-h-gh.dat};
            \addplot [color=magenta, mark=diamond*, dotted, mark options={solid}]
                table [x index=0, y index=1, y error index=2] {figs/accs/tree_lstm-test-h-gm.dat};
            \addplot [color=magenta, mark=diamond*]
                table [x index=0, y index=1, y error index=2] {figs/accs/tree_lstm-test-h-ghm.dat};
            \addplot [color=red, dashed, mark options={solid}, mark size=1pt]
                table [x index=0, y index=1, y error index=2] {figs/accs/bert-test-h-gh.dat};
            \addplot [color=red, dotted, mark options={solid}, mark size=1pt]
                table [x index=0, y index=1, y error index=2] {figs/accs/bert-test-h-gm.dat};
            \addplot [color=red, mark size=1pt]
                table [x index=0, y index=1, y error index=2] {figs/accs/bert-test-h-ghm.dat};
                
            \end{axis}
        \end{tikzpicture}
    \end{subfigure}
    \hfill
    \begin{subfigure}{}
        \begin{tikzpicture}
            \begin{axis}[
                title={\textbf{Machine}},
                xlabel={Training Set Size},
                ylabel={Accuracy (\%)},
                xmin=2000, xmax=21000,
                ymin=0, ymax=100,
                xtick={5000, 10000, 15000, 20000},
                xticklabels={5K, 10K, 15K, 20K},
                legend pos=north west,
                ymajorgrids=true,
                xmajorgrids=true,
            ]
            
            \addplot [color=orange, no marks]
                table [x index=0, y index=1]
                {figs/accs/rnd-hm.dat};
            
            \addplot [color=blue, mark=triangle*, dashed, mark options={solid}]
                table [x index=0, y index=1, y error index=2] {figs/accs/lstm-test-m-gh.dat};
            \addplot [color=blue, mark=triangle*, dotted, mark options={solid}]
                table [x index=0, y index=1, y error index=2] {figs/accs/lstm-test-m-gm.dat};
            \addplot [color=blue, mark=triangle*]
                table [x index=0, y index=1, y error index=2] {figs/accs/lstm-test-m-ghm.dat};
            \addplot [color=magenta, mark=diamond*, dashed, mark options={solid}]
                table [x index=0, y index=1, y error index=2] {figs/accs/tree_lstm-test-m-gh.dat};
            \addplot [color=magenta, mark=diamond*, dotted, mark options={solid}]
                table [x index=0, y index=1, y error index=2] {figs/accs/tree_lstm-test-m-gm.dat};
            \addplot [color=magenta, mark=diamond*]
                table [x index=0, y index=1, y error index=2] {figs/accs/tree_lstm-test-m-ghm.dat};
            \addplot [color=red, dashed, mark options={solid}, mark size=1pt]
                table [x index=0, y index=1, y error index=2] {figs/accs/bert-test-m-gh.dat};
            \addplot [color=red, dotted, mark options={solid}, mark size=1pt]
                table [x index=0, y index=1, y error index=2] {figs/accs/bert-test-m-gm.dat};
            \addplot [color=red, mark size=1pt]
                table [x index=0, y index=1, y error index=2] {figs/accs/bert-test-m-ghm.dat};
                
            \end{axis}
        \end{tikzpicture}
    \end{subfigure}
    
    \begin{subfigure}{}
        \begin{tikzpicture} 
            \begin{axis}[%
            hide axis,
            xmin=0, xmax=1,
            ymin=0, ymax=1,
            legend style={
                draw=white!15!black,
                legend cell align=left,
                legend columns=3,
                column sep=1ex,
            }
            ]
            \addlegendimage{color=blue, mark=triangle*, dashed, mark options={solid}}
            \addlegendentry{LSTM-H};
            \addlegendimage{color=blue, mark=triangle*, dotted, mark options={solid}}
            \addlegendentry{LSTM-M};
            \addlegendimage{color=blue, mark=triangle*}
            \addlegendentry{LSTM-HM};
            \addlegendimage{color=magenta, mark=diamond*, dashed, mark options={solid}}
            \addlegendentry{Tree-LSTM-H};
            \addlegendimage{color=magenta, mark=diamond*, dotted, mark options={solid}}
            \addlegendentry{Tree-LSTM-M};
            \addlegendimage{color=magenta, mark=diamond*}
            \addlegendentry{Tree-LSTM-HM};
            \addlegendimage{color=red, dashed, mark options={solid}, mark size=1pt}
            \addlegendentry{BERT-H};
            \addlegendimage{color=red, dotted, mark options={solid}, mark size=1pt}
            \addlegendentry{BERT-M};
            \addlegendimage{color=red, mark size=1pt}
            \addlegendentry{BERT-HM};
            \addlegendimage{color=orange, no marks};
            \addlegendentry{Random Baseline};
            \end{axis}
        \end{tikzpicture}
    \end{subfigure}
    \caption{Test accuracies of models on different subsets. The suffix of a model indicates the training set it is trained on. We report mean accuracies over 5 runs.}
    \label{fig:classify}
\end{figure*}

\item[Detecting Structures] The subset accuracy reflects the structural difference of translations and a model's ability to detect it. The highest Gold accuracy come from the -M models, probably because it is easier for the model to distinguish Gold and Machine translations due to the broken structures of machine translation, indicated by the high NNP percentage and low CC percentage. LSTM-M has an unexpected low performance on the Machine set, maybe due to its inability to detect structure. In contrast, the Human translations resemble Gold translations in structure more than Machine translations, which may confuse the -H and -HM models. 

\item[Confusion from Structures] We also notice the interesting fact that the Gold accuracy BERT deteriorates more rapidly than Tree-LSTM. This may be because that the Tree-LSTM model is better at explicitly identifying good structures that adds credibility to Gold translations. In the Human set, however, the identification of structure confuses both Tree-LSTM and BERT, preventing them from out-performing the sequential LSTM that is unaware of the structure. The Machine set creates no such confusion. 

\end{description}

A detailed analysis of the failure cases would require linguistic knowledge, which is beyond the scope of this paper. Coarse-grained metrics such as the length or the POS percentage are not sufficient. 

\subsection{Rewriting}

The good performance of -HM models when trained on the full set justified our choice of training set for the rewriting task. The decreasing accuracy on Gold translations has less effect on the rewriting task, since we do not rewrite Gold translations.

Despite the success of classification models, the training of the rewriting model failed. During training, though the perplexity on the training set is reduced, the perplexity on the development set remains high, indicating an inevitable over-fitting. The performance on the test set is visibly poor, defying the need for calculating automatic metrics.

We believe that OpenNMT fails because the rewriting task is too aggressive. The average token-level Levenshtein distance from Human/Machine to Gold is 5.39/4.93, close to the average Gold token length 4.97. Our situation is very different from the traditional sentence simplification tasks, where strategies are much more conservative and the overlapping rate is high \citep{zhang-lapata-2017-sentence}. To rewrite, the model even has too guess from partial translations. The parametric knowledge in GloVe vectors does not suffice, and non-parametric knowledge should be incorporated by methods such as retrieval-augmented generation \citep{lewis2020retrieval}.

\section{Related Work}
\label{section:related-work}

\begin{description}[style=unboxed,leftmargin=0cm, labelsep=1em, itemsep=0em]
    \item[Leveraging Human Translation] There are works that leverage existing human translations, either from experts \citep{shimizu-etal-2014-collection} or language learners \citep{dmitrieva-tiedemann-2021-creating}. The collected datasets are innately more varied than the parallel corpus collected by back-translation or other automated methods, such as in \citep{kim-etal-2021-bisect}. In our work, we try to use both human and machine translations to increase the size of PETCI, and they cooperate well.
    \item[Faster BERT] Ideally, after expanding the dataset, the model needs to be re-trained. The sheer size of Transformer-based models may prohibit language learners from using it. Smaller versions of BERT have been proposed, such as DistilBERT \citep{sanh2020distilbert}. However, the 40\% decrease in model size comes at a cost of 3\% decrease in performance, which may cause BERT to lose its slight advantage over smaller models such as Tree-LSTM. Similarly, quantization and pruning compress the BERT model at a factor of no more than 10, but at a cost of 2\% decrease in performance \citep{cheong2019transformers}.
    \item[Evaluation Metrics] Our rewriting task is close to sentence simplification. Metrics such as BLEU \citep{papineni-etal-2002-bleu} can be used to evaluate sequence generation quality, but are shown to be less correlated with human judgements on simplification quality than another metric SARI \citep{xu-etal-2016-optimizing}. However, to establish the correlation between any of these metrics with PETCI rewriting quality, we need fine-grained labels. The Flesch-Kincaid readability score can also be used to judge the readability of translation, but it is not guaranteed to correlate with the ease of understanding idiom translations. The cosine similarity between translations in PETCI is not calculated, but a low similarity is expected due to the pervasive metaphoric language in idioms. 
\end{description}

\section{Conclusion}
\label{section:conclusion}
In this paper, we build a Parallel English Translation dataset on Chinese Idioms (PETCI) and show its advantages over the existing Chinese idiom datasets. Based on PETCI, we propose a binary classification and a rewriting task, and the baseline models we choose succeed on the former, meanwhile showing great potential when the dataset size increases. Also, we build the datasets and choose the models to make them accessible to language learners, which allows further improvement without the involvement of translation experts. 

\section*{Acknowledgements}

The author thanks Natsume Shiki for insightful comments and kind support.

\bibliography{anthology,custom}
\bibliographystyle{acl_natbib}

\vfill
\pagebreak

\appendix

\begin{figure*}[t]
    \centering
    \begin{subfigure}{}
        \includegraphics[width=14cm]{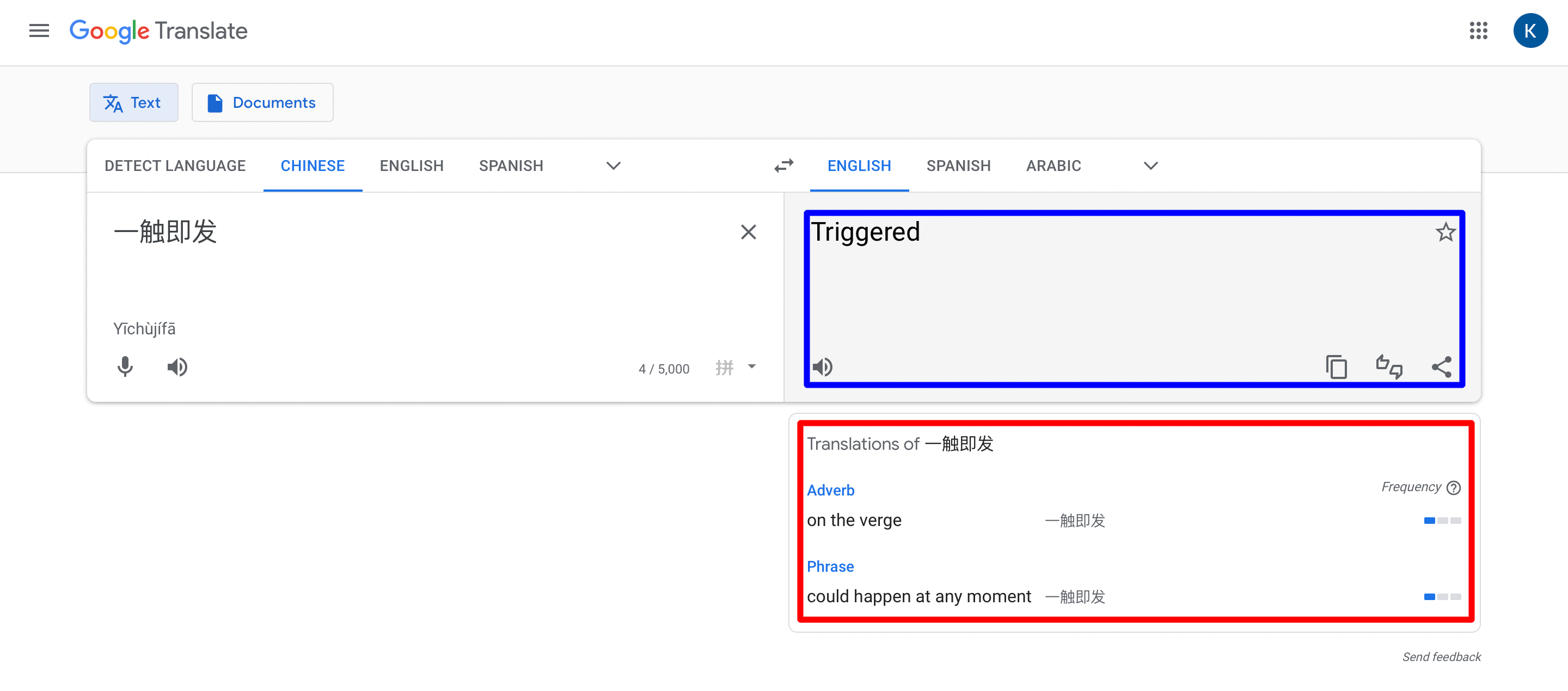}
    \end{subfigure}
    \begin{subfigure}{}
        \includegraphics[width=14cm]{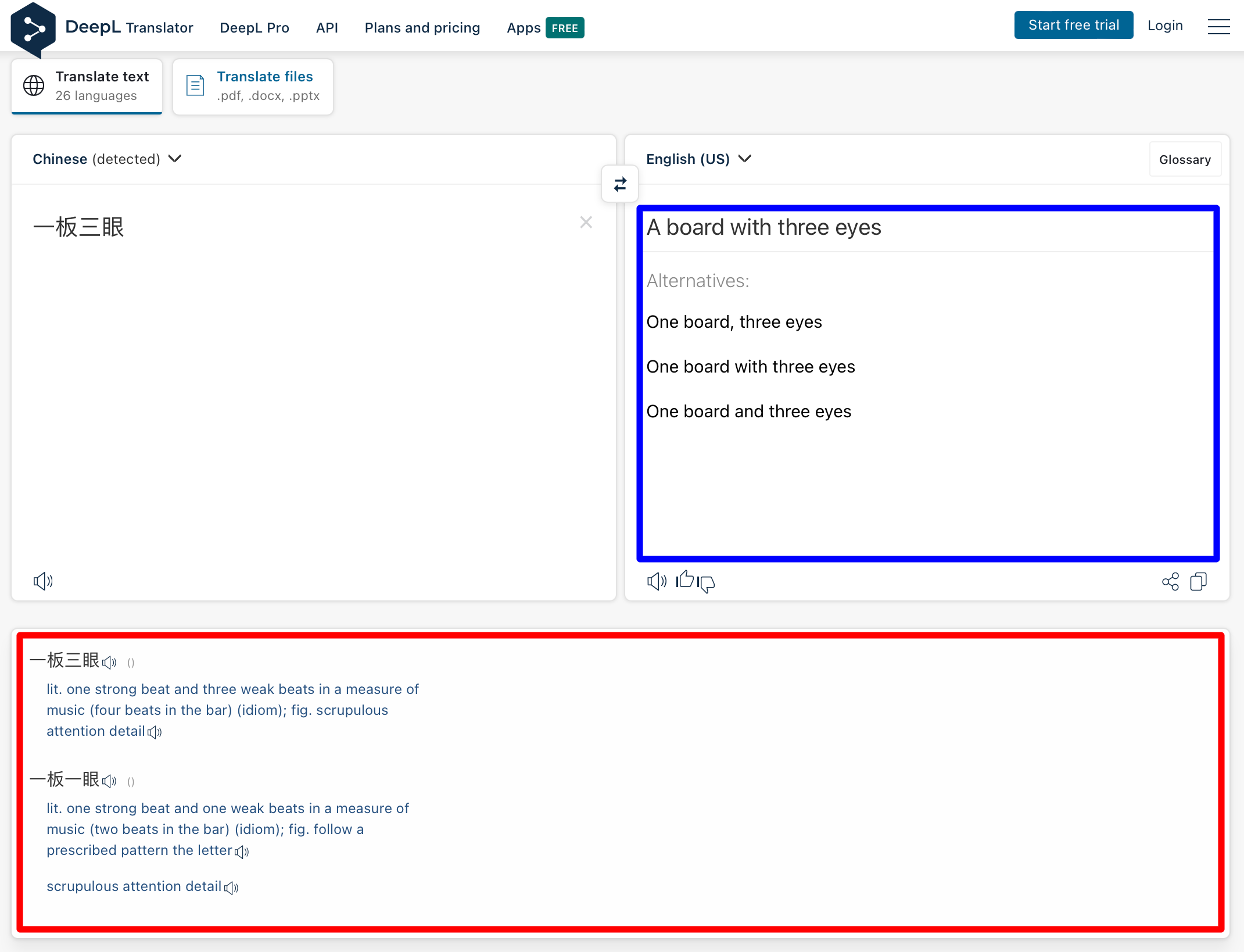}
    \end{subfigure}
    \caption{The web interface of Google (top) and DeepL (bottom) translation. We take screenshots and perform OCR on the region in the blue box. There are other useful information in the red box, but these content are not consistently provided for each idiom, so we ignore them.}
    \label{fig:interface}
\end{figure*}

\section{Data Cleaning Details}
\label{section:cleaning}

It takes around 50 hours to manually clean up and add the Chinese idioms into the OCR results. Special characters replace alphabetic characters in the OCR results of the dictionary, such as @ for a. Also, letters with similar shape are sometimes confused, such as \{a, e, o, s\}, \{i, l, t\}, and \{r, v\}. The results are then checked by the spelling correction provided by the Mac TextEdit. The example sentences are removed.

\section{Web Interface}
\label{section:interface}

The screenshots of Google and DeepL translation web interface are shown in Figure \ref{fig:interface}. The translation data is collected on February 5 and 6 for DeepL and Google, and he screenshots are taken on February 17, 2022. The results are the same across the short time span. Extra information returned from the translation model, such as dictionary translations (shown in the red box), are not used in this paper because they are neither easily scraped nor consistently returned. Also, the alternative results from DeepL are not returned when we put more than one line of text in the source box.

\section{Further Training Details}
\label{section:training-details}
In this section, we list the hyperparameters we have used for all models. If not listed, the hyperparamters are set to the default values provided by the packages.

Hyperparameters for LSTM are:
\begin{itemize}[itemsep=0em]
    \item memory dimension: 168
    \item batch size: 25
    \item dropout rate: 0.5
    \item optimizer: AdaGrad
    \item learning rate: 0.05
\end{itemize}

Hyperparameters for Tree-LSTM are:
\begin{itemize}[itemsep=0em]
    \item memory dimension: 150
    \item batch size: 25
    \item dropout rate: 0.5
    \item optimizer: AdaGrad
    \item learning rate: 0.05
\end{itemize}

For LSTM and Tree-LSTM, we also try a batch size of 20, but observe no qualitatively different results.

Hyperparameters for BERT are:
\begin{itemize}[itemsep=0em]
    \item batch size: 16
    \item epochs: 3
    \item save steps: 500
    \item warmup steps: 500
    \item optimizer: AdamW
    \item learning rate: 5e-5
    \item weight decay: 0.01
\end{itemize}

Hyperparameters for OpenNMT are:
\begin{itemize}[itemsep=0em]
    \item LSTM layers: 2
    \item hidden state size: 500
    \item hidden units: 500
    \item dropout probablity: 0.3
    \item save steps: 300
    \item optimizer: SGD
    \item step when learning rate is halved: 2400
    \item learning rate: 1
\end{itemize}

For OpenNMT, we follow the advice of \citep{cooper-shardlow-2020-combinmt} to convert the number of epochs to steps when re-implementing the model from \citep{nisioi-etal-2017-exploring}. We also try to change the optimizer to AdaGrad or Adam with the recommended learning rate, but the training does not succeed.

For LSTM and Tree-LSTM, we apply early stopping when the accuracy does not increase on the development set for 10 consecutive epochs. For BERT, we choose from the last 5 checkpoints the checkpoint that has the best accuracy on the development set. For OpenNMT, we apply early stopping when perplexity has not improved on the development set for 5 checkpoints.

Due to the drastic differences in the number of parameters (Table \ref{tab:num-par}), the LSTM and Tree-LSTM model are trained on CPU, while the BERT and OpenNMT model are trained on 1 GPU assigned by Google Colab.

\begin{table}
\centering
\begin{tabular}{cc}
\hline
\textbf{Model} & $|\theta|$\\
\hline
LSTM & 316k \\
Tree-LSTM & 316k \\ 
BERT & 110M \\
OpenNMT & 23M \\
\hline
\end{tabular}
\caption{Number of parameters.}
\label{tab:num-par}
\end{table}

\section{Numeric Results}
\label{section:numeric}

Table \ref{tab:numeric} shows the numeric results of Figure \ref{fig:classify}.

\begin{table*}[]
    \centering
    \addtolength{\leftskip} {-2cm}
    \addtolength{\rightskip}{-2cm}
    \begin{tabular}{c|l|ccccc}
        \hline
        \multirow{2}{*}{\textbf{Test Set}} & \hfil\multirow{2}{*}{\textbf{Model}}\hfill & \multicolumn{5}{c}{\textbf{Training Set Size}}\\ \cline{3-7}
        & & 1 & 2 & 3 & 4 & 5 \\ \hline
        \multirow{9}{*}{Overall} & LSTM-H & 54.75 (1.92) & 67.84 (3.26) & 70.75 (3.85) & 73.61 (2.32) & 75.66 (1.75) \\
        & LSTM-M & 36.83 (1.93) & 40.67 (1.54) & 45.21 (1.83) & 45.93 (2.73) & 43.65 (3.57)\\
        & LSTM-HM & 62.55 (1.20) & 70.47 (2.18) & 69.69 (3.09) & 72.52 (1.14) & 73.53 (2.05) \\
        & Tree-LSTM-H & 52.34 (5.31) & 52.66 (6.06) & 54.16 (1.97) & 56.45 (0.55) & 64.31 (2.10) \\
        & Tree-LSTM-M & 53.57 (1.23) & 60.72 (1.10) & 62.52 (1.54) & 64.06 (1.45) & 66.43 (1.74) \\
        & Tree-LSTM-HM & 64.50 (1.53) & 70.30 (1.21) & 73.82 (1.24) & 75.71 (1.42) & 76.84 (1.44) \\
        & BERT-H & 42.41 (2.18) & 58.09 (7.47) & 60.15 (3.41) & 62.14 (2.29) & 63.57 (2.45) \\
        & BERT-M & 55.01 (5.92) & 58.33 (4.07) & 64.24 (0.65) & 64.72 (1.64) & 67.28 (1.97) \\
        & BERT-HM & 62.60 (3.25) & 71.59 (1.34) & 75.88 (0.79) & 77.19 (0.91) & 78.56 (0.36) \\ \hline
        \multirow{9}{*}{Gold} & LSTM-H & 29.14 (2.59) & 15.27 (4.10) & 14.48 (4.58) & 10.81 (3.07) & 9.65 (1.48) \\
        & LSTM-M & 78.38 (2.55) & 71.32 (2.55) & 63.57 (3.09) & 64.96 (3.00) & 67.66 (4.79) \\
        & LSTM-HM & 31.23 (6.44) & 19.07 (3.25) & 21.81 (4.85) & 19.77 (1.33) & 16.94 (3.01) \\
        & Tree-LSTM-H & 36.24 (6.63) & 37.68 (11.12) & 36.61 (5.18) & 36.98 (0.95) & 32.34 (3.50) \\
        & Tree-LSTM-M & 73.88 (2.56) & 66.68 (1.42) & 63.43 (3.10) & 61.67 (2.45) & 57.45 (3.34) \\
        & Tree-LSTM-HM & 48.45 (1.67) & 37.08 (1.58) & 33.32 (1.19) & 28.72 (2.06) & 27.33 (2.12) \\
        & BERT-H & 65.90 (3.95) & 38.79 (11.03) & 37.68 (10.55) & 37.31 (5.10) & 38.61 (5.69) \\
        & BERT-M & 74.85 (7.73) & 71.28 (6.61) & 64.83 (1.87) & 62.37 (3.70) & 59.12 (2.46) \\
        & BERT-HM & 59.54 (4.05) & 42.32 (3.82) & 34.85 (2.54) & 29.79 (2.32) & 25.99 (1.18) \\ \hline
        \multirow{9}{*}{Human} & LSTM-H & 69.62 (2.31) & 84.09 (3.31) & 85.65 (4.11) & 88.74 (2.33) & 90.85 (2.16) \\
        & LSTM-M & 23.07 (2.28) & 28.87 (2.54) & 35.20 (2.53) & 35.09 (3.35) & 32.32 (4.26) \\
        & LSTM-HM & 69.71 (2.72) & 80.90 (4.10) & 77.73 (5.56) & 80.73 (2.12) & 83.03 (2.87) \\
        & Tree-LSTM-H & 63.37 (6.04) & 63.59 (8.51) & 65.64 (2.96) & 65.95 (0.75) & 73.06 (1.97) \\
        & Tree-LSTM-M & 30.36 (1.46) & 37.42 (0.90) & 40.50 (2.78) & 43.55 (2.50) & 49.45 (3.30) \\
        & Tree-LSTM-HM & 58.36 (2.02) & 67.87 (1.97) & 73.97 (1.69) & 78.03 (2.25) & 80.11 (2.69) \\
        & BERT-H & 45.12 (3.92) & 70.57 (10.04) & 72.47 (6.51) & 73.73 (3.43) & 75.00 (4.90) \\
        & BERT-M & 28.65 (8.48) & 31.27 (6.19) & 39.72 (1.50) & 41.90 (3.30) & 46.29 (4.28) \\
        & BERT-HM & 51.33 (4.64) & 68.02 (2.34) & 76.19 (1.80) & 79.98 (2.64) & 83.09 (0.46) \\ \hline
        \multirow{9}{*}{Machine} & LSTM-H & 51.17 (2.93) & 71.58 (5.59) & 76.72 (6.23) & 81.40 (4.00) & 84.42 (2.46) \\
        & LSTM-M & 34.59 (3.02) & 40.32 (2.07) & 47.29 (2.79) & 48.43 (4.08) & 44.99 (5.66) \\
        & LSTM-HM & 66.72 (2.99) & 78.38 (2.39) & 78.37 (3.59) & 82.60 (1.71) & 83.76 (3.13) \\
        & Tree-LSTM-H & 48.79 (8.37) & 48.82 (9.51) & 50.72 (3.26) & 55.14 (0.93) & 67.48 (3.51) \\
        & Tree-LSTM-M & 65.28 (2.22) & 76.98 (1.62) & 79.35 (1.66) & 80.74 (1.73) & 82.42 (2.25) \\
        & Tree-LSTM-HM & 74.27 (2.11) & 82.55 (1.67) & 86.32 (1.63) & 88.55 (1.69) & 89.73 (1.59) \\
        & BERT-H & 32.99 (2.97) & 54.41 (11.31) & 57.59 (5.14) & 60.87 (3.38) & 62.47 (3.63) \\
        & BERT-M & 69.31 (8.18) & 75.33 (5.60) & 83.12 (0.71) & 83.20 (2.05) & 86.15 (1.48) \\
        & BERT-HM & 72.31 (4.45) & 83.48 (2.15) & 88.43 (1.32) & 89.79 (0.63) & 91.42 (0.72) \\ \hline
    \end{tabular}
    \caption{The classification accuracies on PETCI. We report mean accuracy over 5 runs (standard deviation in parentheses). The training sets are labelled by their relative sizes, with 1 being the smallest training set, and 5 being the full training set. The suffix of model indicates the type of training set on which the model is trained.}
    \label{tab:numeric}
\end{table*}

\end{document}